\documentclass[runningheads]{llncs}
\usepackage[T1]{fontenc}
\usepackage{graphicx}
\usepackage{booktabs}
\usepackage[misc]{ifsym}
\usepackage{mwe}
\usepackage{subcaption}
\usepackage{float}
\usepackage{enumitem}
\usepackage{xcolor}
\usepackage{amsmath}
\definecolor{linkblue}{RGB}{0,90,181}
\usepackage[colorlinks=true, linkcolor=linkblue, urlcolor=linkblue, citecolor=linkblue]{hyperref}
\usepackage{orcidlink}
\usepackage{tabularx}
\usepackage{booktabs}
\usepackage{ragged2e} 
\usepackage{makecell}
\usepackage{todonotes}
\usepackage{times}

\usepackage{booktabs}
\usepackage{subcaption}
\usepackage{array}
\usepackage{adjustbox}

\newcommand{\corr}{(\Letter)}

\title{A Universal Dense Football Event Representation Based on TabTransformer}
\titlerunning{A Universal Dense Football Event Representation Based on TabTransformer}
\begin{document}



\author{
Weiran Yang\orcidlink{0009-0008-8810-5400}\inst{1} \corr \and
Daniel Memmert\orcidlink{0000-0002-3406-9175}\inst{1} \and
Maximilian Klemp-Weins\orcidlink{0000-0003-4466-8136}\inst{1}
}

\authorrunning{Yang et al.}

\institute{
\leavevmode\inst{1} Institute of Exercise Training and Sport Informatics, German Sport University Cologne, Cologne, Germany\\
\email{apinecdqz@gmail.com,\{d.memmert, m.klemp-weins\}@dshs-koeln.de}\\[0.3em]
}

\maketitle              

\begin{abstract}
Football event data constitute a rich spatiotemporal source for quantitative analysis of player actions in team sports. These datasets contain heterogeneous features, combining continuous location coordinates with categorical variables such as action type, action outcome, and body part. Such data have been applied in sports analytics for match outcome forecasting, player evaluation, and tactical pattern recognition. However, existing approaches predominantly encode categorical features using one-hot or ordinal embedding representations, overlooking the intrinsic semantics of action descriptors. The Transformer is a deep neural network architecture based on self-attention that captures dependencies between input features at arbitrary positions. We propose and implement a Transformer-based model to learn latent dependencies among categorical event features and produce dense representations of football events. By encoding categorical features as learned embedding vectors, sport-specific action semantics are captured during pretraining, enabling the representations to support downstream tasks such as action value estimation and play style recognition. Empirical evaluation shows that the embedding representations yield superior probability calibration over task-specific baselines on the downstream prediction tasks, as measured by Brier score.
\end{abstract}

\newpage{}

\section{Introduction}
Football event data are high-dimensional spatiotemporal records of player actions during a match \cite{gudmundsson2017spatio}. They encode player identities, event types, timestamps, and positional coordinates, forming chronologically ordered event sequences that support a variety of football analysis tasks, including game outcome forecasting \cite{klemp2021b} and player performance analysis \cite{pappalardo2019}. Machine learning methods endow event data analytics with predictive capacity: action value models estimate the contribution of an event to successful outcomes in the near future, based on the event's features and context \cite{decroos2019}, while unsupervised methods have enabled playing style recognition from event streams of individual players or entire teams \cite{decroos2020}. These data-driven analyses have provided insights into player and team evaluation from technical and tactical perspectives, enabling data-driven decision-making at professional clubs.

However, despite these advances, current event data analytics faces two challenges.
First, most analysis tasks require training of separate models with task-specific features, often involving complex manual feature engineering. This increases computational overhead and renders models less generalizable for other tasks.

Second, existing approaches have not fully exploited the intrinsic semantics of features in football event analytics. Feeding spatiotemporal data including categorical features into ML pipelines requires some kind of representation that is usable for the models \cite{raabe2023}, and current approaches have not sufficiently acknowledged the semantic dependencies among feature levels: standard categorical encodings (one-hot and ordinal) project each action type to an arbitrary numerical code, discarding relational semantics. For example, the representational similarity between a \textit{cross} and a \textit{through ball} is no different from the similarity between a \textit{cross} and a \textit{tackle}, even though the former pair shares the spatial and tactical context of a final-third ball delivery \cite{borisov2021}.

Transformer architectures \cite{vaswani2017} offer a potential remedy to both challenges through their capacity for transferable representation learning and contextual categorical embedding. A central advantage of Transformer-based models is their ability to learn dense embeddings during large-scale pretraining that can subsequently be transferred across multiple downstream tasks with minimal task-specific adaptation. 

Despite the growing adoption of Transformer architectures across many domains, their application in football event analytics remains limited. To our knowledge, the only published Transformer-based framework for football event analysis \cite{10.1007/978-3-031-86692-0_7} represents each event as a single atomic token within a sequential modeling setup and retains player and team identifiers during training. 
As discussed in Section 2, this design prevents feature-level interaction modeling and risks conflating action quality with player reputation, limiting transferability across competitions and leagues. Furthermore, the framework is optimized for a single downstream objective and does not address the broader challenge of learning transferable event representations that can generalize across multiple football analytics tasks.

To address these limitations, we propose a TabTransformer-based framework \cite{huang2020b} for learning universal dense representations of football events through feature-level self-supervised pretraining on anonymized event data. TabTransformer applies self-attention directly to categorical feature embeddings while integrating continuous variables through conventional feedforward layers. By contextualizing categorical features at the feature level rather than treating entire records as indivisible tokens, TabTransformer learns interactions between variables within a structured tabular representation.

We introduce an anonymized Masked Language Modeling pretraining protocol that captures dependencies among event features without encoding player- or team-specific biases. The resulting event embeddings can subsequently be reused across distinct downstream tasks without requiring separate task-specific representation learning or extensive manual feature engineering. We evaluate the learned representations on three downstream tasks: Expected Goals (xG), VAEP scoring and conceding prediction, and player style evaluation using public professional football event data from StatsBomb \cite{statsbomb2024}. 

Experimental results demonstrate that the proposed framework achieves competitive predictive performance across two predictive downstream tasks while relying on a single transferable event representation. The framework is further used on a third, unsupervised downstream task, retrieving players in a real-world scouting application using similarity search on their averaged event embeddings.

\section{Related Work}

\subsection{Data-Driven Football Event Data Analytics}
Football event data constitute rich spatiotemporal records generated during matches. Beyond player identity, location coordinates, and timestamps, event records encode action-specific attributes such as the body part used and the action outcome. Together, these features provide a detailed description of on-pitch actions and the progression of play. Aggregated across a match, event data form a heterogeneous tabular dataset comprising continuous features (e.g., spatial coordinates) alongside categorical features (e.g., event type, player identity).

On average, approximately 1{,}700 on-ball events are recorded per match in professional football \cite{pappalardo2019b}. Compared to high-frequency tracking data, event data are more amenable to manual annotation and collection. Several commercial providers, including Opta Stats Perform, Hudl StatsBomb, and WyScout, have built large football event databases with heterogeneous schemas. These databases enable professional clubs to quantitatively evaluate player performance, diagnose tactical execution, and identify prospective players under a data-driven framework.

\subsection{Machine Learning Methods in Data-Driven Football Analytics}
Machine learning methods have been applied to identify key actions from the large volumes of event data generated per match \cite{tuyls2021}. Predictive performance indicators such as Expected Goals (xG) \cite{DBLP:conf/pkdd/RobberechtsD20}, Expected Assists (xA), and the Value of Actions by Estimating Probabilities (VAEP) \cite{decroos2019} are derived via supervised learning on labeled event data. Due to the prevalence of categorical features in event data, gradient-boosted decision tree (GBDT) models such as XGBoost and CatBoost dominate event analytics, owing to their native handling of mixed-type tabular data \cite{klemp2026,biermann2024quantification}.

Deep learning methods have also attracted attention in data-driven football event analytics. Recurrent architectures such as Long Short-Term Memory (LSTM) networks \cite{hochreiter1997long} and Gated Recurrent Units (GRU) \cite{DBLP:conf/emnlp/ChoMGBBSB14} have been used to model temporal dependencies across event sequences. Merhej et al.\ \cite{merhej2021} proposed a deep learning framework for valuing defensive actions in football event data, demonstrating that neural architectures can capture contextual dependencies that tree-based models miss. Convolutional Neural Networks (CNNs), effective at processing spatial representations, have been applied to football analytics for evaluating player positioning decisions \cite{wagenaar2017}; however, the sparse distribution of player coordinates over pitch space makes training challenging. Graph Neural Networks (GNNs) have demonstrated efficiency on spatiotemporal sports data by incorporating the topological relationships between players based on their distances and interactions \cite{raabe2023}. Although successfully applied to tracking data, GNNs incur a high computational cost when applied to tabular event data.

\subsection{Transformer and Attention Mechanism}
Transformers \cite{vaswani2017} have been widely applied beyond natural language processing to spatiotemporal and structured data.
A key driver of this transfer is the Masked Language Modeling (MLM) pretraining objective, introduced in BERT \cite{devlin2018}, in which a fraction of input tokens is randomly masked and the model is trained to reconstruct them from surrounding context, enforcing bidirectional context awareness and yielding transferable representations that can be fine-tuned efficiently with limited labeled data.

For heterogeneous tabular data, TabTransformer \cite{huang2020b} demonstrated that applying self-attention specifically to categorical feature embeddings yields robust contextual representations that outperform standard deep learning approaches and match gradient-boosted tree ensembles.
Subsequent work including SAINT \cite{somepalli2021} and FT-Transformer \cite{gorishniy2021} further established Transformer-based architectures as viable alternatives to tree-based models for tabular data \cite{borisov2021}, motivating their adoption for football event data, which exhibits the same heterogeneous tabular structure.

The most closely related approach to our work is SoccerTransformer \cite{10.1007/978-3-031-86692-0_7}, which employs self-supervised pretraining on football event data for attack outcome classification and player rating.
However, SoccerTransformer operates at the event-sequence level, treating each event as an atomic token, and retains player and team identifiers during training.
Our approach differs in three respects: (1) self-attention is applied across feature columns \emph{within} a single event rather than across event tokens, enabling explicit modeling of interactions between individual event attributes; (2) player and team identifiers are removed prior to pretraining to prevent ability-based confounding, improving transferability across competitions; and (3) evaluation covers VAEP in addition to goal prediction, demonstrating generalization across structurally distinct downstream tasks.

\subsection{Data Representation in Football Event Data Analytics}
(1) \textit{One-hot encoding.} Machine learning methods cannot directly process categorical features stored as character strings. A common approach is to encode each category as a binary vector of length equal to the number of categories $N$, with a single element set to one and all others to zero. While this renders categorical data numerically tractable, it treats all categories as equidistant, ignoring semantic similarities between related actions such as crosses and passes. High-cardinality features such as player identifiers also lead to very high-dimensional and sparse representations, increasing the computational burden for tree-based models.

(2) \textit{Ordinal encoding.} Ordinal encoding maps each category to a unique integer, requiring only a single value per feature and thus avoiding the dimensionality problem of one-hot encoding. However, it similarly fails to capture semantic relationships between categories, as the assigned integers carry no meaningful ordering.

(3) \textit{Dense vector representation.} To address the semantic limitations of both approaches, researchers have explored dense, continuous representations. Autoregressive language models and embedding-based methods generate low-dimensional vectors, mapping semantically related inputs to nearby points in the representation space.

(4) \textit{Vector Representation in Player Style Recognition:}
Inspired by word embedding methods such as Word2Vec \cite{mikolov2013} and Doc2Vec \cite{le2014}, which learn distributed representations by predicting contextual co-occurrences in text, researchers have applied analogous techniques to football event streams.
Decroos and Davis \cite{decroos2020b} proposed \emph{Player Vectors}, which characterize a player's style by a distribution over action types and outcomes derived from their match event history, yielding low-dimensional vectors suitable for scouting and similarity retrieval.
While Player Vectors aggregate event statistics at the player level, our approach produces a dense representation for each individual event, capturing the contextual interaction of its features rather than the statistical tendencies of the player who performed it, which enables the embedding to serve as input for arbitrary downstream models without per-player data aggregation.

\section{Method}
We propose a pipeline for training a TabTransformer to produce dense event embeddings suitable for multiple downstream tasks.

\subsection{Model Architecture}
\begin{figure}[t]
   \centering
     \includegraphics[width=1\textwidth]{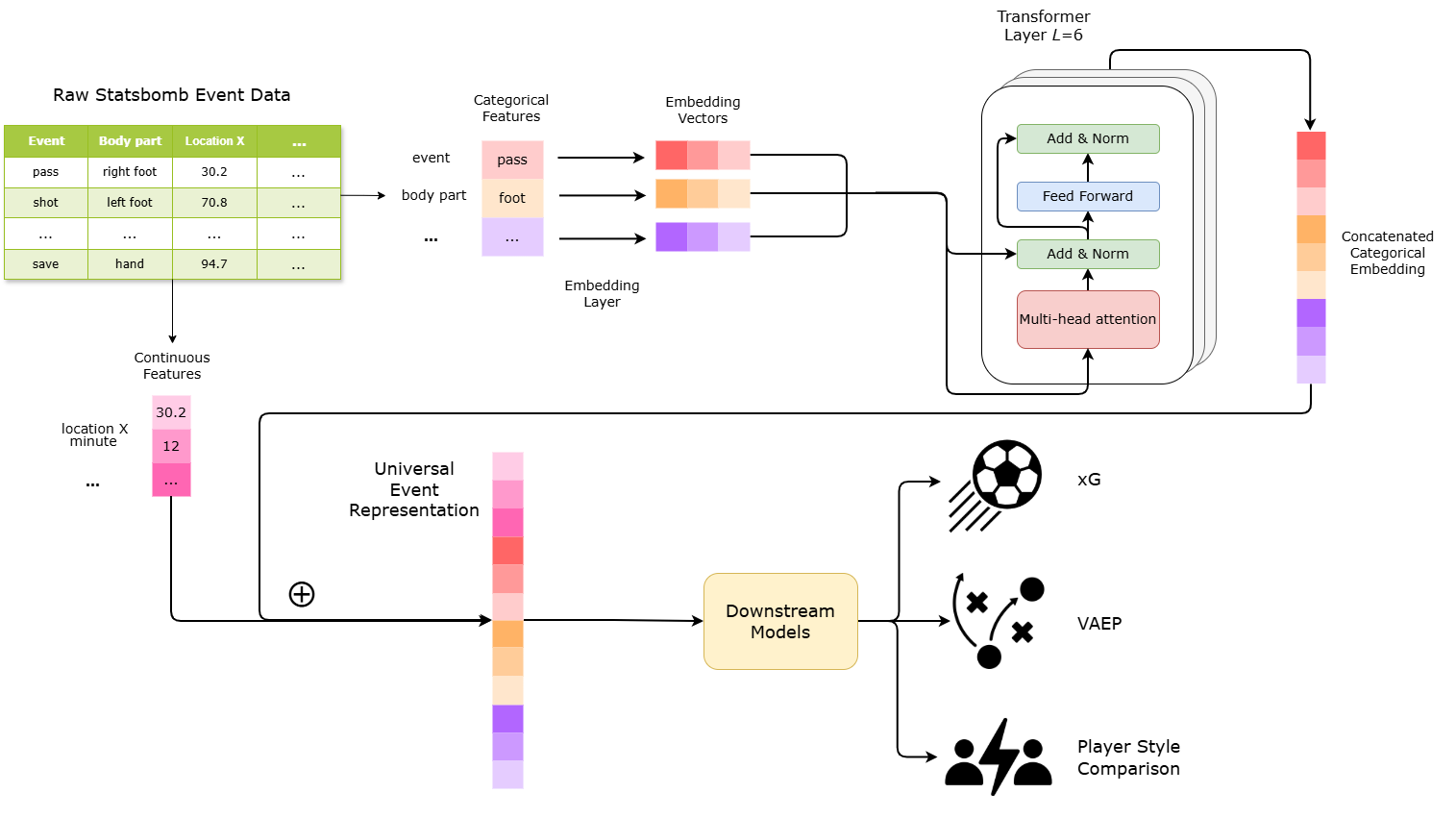}
     \caption{Overview of the proposed framework. Categorical event features are mapped to learned embeddings and contextualized via Transformer self-attention; the resulting 911-dimensional dense representation is reused without retraining across the three downstream evaluation tasks.}
     \label{fig:tabtransformer}
\end{figure}

The proposed model follows the TabTransformer architecture \cite{huang2020b}.
Each categorical feature in the preprocessed event is mapped to a learned $d$-dimensional embedding vector via a dedicated embedding layer.
Continuous features (e.g., event location coordinates, timestamps) are kept as continuous scalars.
The sequence of categorical embeddings is processed by a stack of $L$ Transformer encoder layers, each comprising multi-head self-attention and a position-wise feed-forward network with layer normalization.
Self-attention across the feature embeddings allows the model to capture pairwise dependencies between features; for example, it can learn that a right-footed cross from a wide position constitutes a semantically coherent combination.
The contextualized output embeddings of all $K$ categorical features are concatenated with the $d_{num}$ raw numerical features to form the final event representation vector.
For an event with $K$ categorical features each embedded to dimension $d$, and $d_{\text{con}}$ continuous features, the output dimensionality is $K \cdot d + d_{\text{con}}$.
In our implementation using StatsBomb open data \cite{statsbomb2024}, this yields a 911-dimensional event embedding. The key architecture is illustrated in Figure \ref{fig:tabtransformer}, and the pretraining hyperparameters are summarized in Table~\ref{tab:hyperparams}.

\begin{table}[t]
\vspace{-1.5em}
\centering
\scriptsize 
\caption{Model pretraining hyperparameters}
\label{tab:hyperparams}

\begin{tabular}{l@{\hspace{2em}}l}
\toprule
Hyperparameter & Value \\
\midrule
Transformer layers ($L$)  & 6 \\
Attention heads           & 8 \\
Embedding dimension ($d$) & 32 \\
Masking rate              & 0.15 \\
\bottomrule
\end{tabular}
\hspace{3em} 
\begin{tabular}{l@{\hspace{2em}}l}
\toprule
Hyperparameter & Value \\
\midrule
Optimizer       & Adam \\
Learning rate   & 1e-3 \\
Batch size      & 256 \\
Training epochs & 20 \\
\bottomrule
\end{tabular}
\vspace{-1.5em}
\end{table}

\subsection{Dataset}
The StatsBomb open data \cite{statsbomb2024} is used for all experiments. The dataset contains event records from professional football matches, including spatial coordinates, timestamps, event types, and contextual attributes. The training set comprises 6,391,338 events from 1,823 matches in the five major European professional leagues (English Premier League, La Liga, Bundesliga, Serie A, and Ligue 1) during the 2015/16 season. For evaluation, we held out the event datasets of UEFA Euro 2020 (192,692 events in 51 matches), FIFA World Cup 2022 (234,652 events in 64 matches), and Copa America 2024 (100,305 events in 32 matches) as the test set.

\subsection{Pretraining of TabTransformer}
In the preprocessing stage, player and team identifiers are removed to prevent the model from learning player- or team-specific biases. Additionally, posterior features such as shot end coordinates are excluded, as these encode the action outcome and would constitute a form of data leakage during pretraining for relevant downstream tasks.

In order to capture the spatial characteristics of actions, geometric features such as spatial displacement, the angle to goal, and the distance to goal from the event location are derived from the raw data. Additionally, event locations are discretized into a 16 × 12 pitch grid to address the sparsity of continuous spatial coordinates, yielding two additional categorical features (start zone, end zone).

After preprocessing, 28 categorical features and 15 continuous features are preserved.
To capture dependencies among the categorical features of each event, we employed the Masked Language Modeling (MLM) objective \cite{devlin2018} to pretrain the TabTransformer in a self-supervised manner. Following the MLM paradigm, each categorical feature is independently masked with probability 0.15, and the model is trained to reconstruct the masked features from the remaining features of the same event. This objective requires the model to develop a holistic understanding of all input features, making it well-suited for self-attention-based architectures. As a result, the model learns the joint semantics of feature combinations, such as the association between a right-footed action and a position in the attacking third, or a long pass originating from a defensive zone.

\section{Evaluation}
To evaluate the representational capacity of the event embeddings produced by the pretrained TabTransformer, we conduct experiments on two downstream prediction tasks: Expected Goals (xG) and the Value of Actions by Estimating Probabilities (VAEP).
\subsection{Downstream Task 1: Expected Goals (xG)}
Expected Goals (xG) is a widely used metric in football analytics that quantifies the probability of scoring from a shot \cite{DBLP:conf/pkdd/RobberechtsD20}. It reflects the quality of a scoring opportunity at the moment of the shot, providing an informative signal even when no goal is scored. Due to the broad definition of xG, a variety of predictive models can be applied using heterogeneous input features. A basic xG model predicts shot outcome from features such as shot location and the body part used. To incorporate richer contextual information from the events preceding the shot, a sequence of actions can be modelled with sequence architectures such as Long Short-Term Memory (LSTM) networks or Gated Recurrent Units (GRU).

In this experiment, the event datasets were segmented into attacking sequences, defined as loosely-bounded continuous periods of possession (no more than two continuous opponent touches) ending with a shot, a foul, a turnover, or the ball going out of play. To reduce computational cost, sequences exceeding 48 events were truncated, and sequences shorter than 48 events were zero-padded to the fixed length, with a binary validity mask to exclude padding positions from attention computations. Open play sequences ending with a shot, including 17,034 in the training dataset and 1,230 in the test dataset, were retained for the xG prediction task. Finally, in the training dataset, the goal/no-goal ratio is 1:8.32, while in the test dataset the ratio is 1:7.72. The training datasets are further split into training and validation data at a 9:1 ratio, enabling best checkpoint selection for final evaluation on the test set. In training, each event was encoded by the pretrained TabTransformer, transforming the raw 44-dimensional StatsBomb feature vector into the 911-dimensional event embedding. A GRU then processed the sequence of embeddings in temporal order to predict the shot outcome.

Since xG is fundamentally a probability estimate, calibration metrics are the primary criterion for model quality \cite{robberechts2020}: a well-calibrated model assigns a probability of 0.2 to shots that score 20\% of the time, regardless of where a fixed decision threshold falls. Accordingly, we evaluate models primarily by Brier score and LogLoss, with F1 and ROC-AUC reported for completeness. StatsBomb provides a proprietary xG model alongside the open data, which serves as an additional reference. The results are shown in Table~\ref{tab:overall_xg_results}.
The TabTransformer achieves a lower Brier score (0.0923) and LogLoss (0.3173) than the MLP baseline (0.1010 and 0.3283 respectively), demonstrating superior calibration. The StatsBomb built-in xG model achieves the best overall calibration, which is expected given its larger proprietary training dataset and additional spatial features capturing player and opponent positioning.

\begin{table}[t]
\vspace{-1.3em}
\centering
\scriptsize
\caption{Performance comparison of xG results}
\label{tab:overall_xg_results}
\begin{tabular}{lccccc}
\toprule
Model & F1 & ROC-AUC & LogLoss & Brier & Threshold \\
\midrule
MLP baseline          & 0.4571 & 0.8142 & 0.3283 & 0.1010 & 0.45 \\
TabTransformer xG     & 0.4211 & 0.7806 & 0.3173 & 0.0923 & 0.35 \\
StatsBomb built-in xG & 0.2456 & 0.8074 & 0.2810 & 0.0806 & 0.50 \\
\bottomrule
\end{tabular}
\vspace{-1.4em}
\end{table}

\subsection{Downstream Task 2: VAEP}
VAEP (Value of Actions by Estimating Probabilities) is an action valuation framework \cite{DBLP:conf/kdd/DecroosBHD19} that assigns a value score to each on-ball action equal to the change it induces in the probability of scoring or conceding within the next $k{=}10$ actions. This provides a quantitative measure of each player's contribution in both attack and defense. A critical comparison of VAEP against the alternative expected threat (xT) framework is provided by Van Roy et al.\ \cite{vanroy2020}, who highlight systematic differences in how the two approaches value specific actions. In practice, VAEP is commonly implemented using gradient-boosted decision tree models such as XGBoost and CatBoost.

In this experiment, a CatBoost classifier was trained on the event embeddings produced by the pretrained TabTransformer, with Principal Component Analysis (PCA) applied to reduce the embeddings to 128 components for computational efficiency. Baseline models using XGBoost and CatBoost, as well as an MLP, were trained and evaluated on the same feature set. The results for the scoring and conceding sub-tasks are shown in Tables~\ref{tab:vaep_test_score} and \ref{tab:vaep_test_concede} respectively, alongside the original VAEP results reported on the SPADL event format \cite{DBLP:conf/kdd/DecroosBHD19}.

\begin{table*}[t]
\vspace{-0.75em}
\centering
\caption{VAEP test set results}
\label{tab:vaep_test_results}
\vspace{-1.4em}
\begin{subtable}[t]{0.48\textwidth}
\centering
\caption{Score task}
\label{tab:vaep_test_score}
\resizebox{\linewidth}{!}{
\begin{tabular}{l l c c}
\toprule
Model & Features & AUC & Brier \\
\midrule
SPADL + XGBoost                  & SPADL (Wyscout)     & 0.756 & 0.014 \\
SPADL + CatBoost                 & SPADL (Wyscout)     & 0.769 & 0.014 \\
SPADL + XGBoost                            & SPADL 27-dim        & 0.730  & 0.182   \\
SPADL + CatBoost                           & SPADL 27-dim        & 0.760  & 0.176   \\
TabTransformer + MLP                       & StatsBomb 44-dim    & 0.827  & 0.011  \\
TabTransformer + CatBoost (PCA-128)        & StatsBomb 44-dim    & 0.830  & 0.126   \\
Baseline MLP                               & StatsBomb 44-dim    & 0.841  & 0.011  \\
CatBoost                                   & StatsBomb 44-dim    & 0.844  & 0.130   \\
XGBoost                                    & StatsBomb 44-dim    & 0.845  & 0.133   \\
\bottomrule
\end{tabular}
} 
\end{subtable}\hfill
%
\begin{subtable}[t]{0.48\textwidth}
\centering
\caption{Concede task}
\label{tab:vaep_test_concede}
\resizebox{\linewidth}{!}{
\begin{tabular}{l l c c}
\toprule
Model & Features & AUC & Brier \\
\midrule
SPADL + XGBoost                 & SPADL (Wyscout)     & 0.726 & 0.006 \\
SPADL + CatBoost                 & SPADL (Wyscout)     & 0.731 & 0.005 \\
SPADL + XGBoost                            & SPADL 27-dim        & 0.751  & 0.159   \\
SPADL + CatBoost                           & SPADL 27-dim        & 0.814  & 0.144   \\
Baseline MLP                               & StatsBomb 44-dim    & 0.815  & 0.006  \\
TabTransformer + CatBoost (PCA-128)        & StatsBomb 44-dim    & 0.832  & 0.078   \\
XGBoost                                    & StatsBomb 44-dim    & 0.869  & 0.090   \\
TabTransformer + MLP                       & StatsBomb 44-dim    & 0.879  & 0.003  \\
CatBoost                                   & StatsBomb 44-dim    & 0.885  & 0.084   \\
\bottomrule
\end{tabular}
} 
\end{subtable}
\vspace{-2.5em}
\end{table*}

\subsection{Downstream Task 3: Player Style and Similarity Search}

To illustrate the value of the semantic structure of the learned representations, we apply the embeddings for a real-world scouting task: finding players who exhibit a playing style similar to certain target players. As a player is leaving a club or retiring, managers might be interested in finding a replacement that resembles them as closely as possible in terms of style and tactical fit. We therefore use our Transformer architecture to identify the 5 closest matches for various elite players on the main positions in our dataset.

For each player in the training dataset, all actions were aggregated by a 2-level average pooling strategy: match-level first, followed by a season-level average across all matches in the 2015/16 season. Player similarity is then assessed as the cosine similarity between the resulting player-level embedding vectors.

Table \ref{tab:elite6_top5} shows the 5 most similar players for six elite players across six positions: goalkeeper, central defender, full back, central midfielder, wide midfielder, and striker. The retrieved lists demonstrate positional coherence: in every case, the top matches play the same position as the query player. The retrieved player names further validate qualitative plausibility—for example, Lewandowski's top matches include Luis Suárez, Zlatan Ibrahimović, Diego Costa, and Karim Benzema, all elite forwards with comparable all-round profiles combining finishing, hold-up play, and link-up ability. The results for Philipp Lahm further illustrate positional generalization: the top matches include Florenzi and Danilo, both known for operating on either flank, consistent with Lahm's own positional versatility. The uniformly high cosine similarity scores (many at 1.0) are a direct consequence of season-level average pooling, which compresses within-position variance and yields highly concentrated player-level representations; future work will investigate whether within-position discriminability can be further improved through contrastive or fine-grained pretraining objectives.

\newcolumntype{L}[1]{>{\raggedright\arraybackslash}p{#1}}

\begin{table*}[htbp]
\vspace{-2.8em}
\centering
\scriptsize
\setlength{\tabcolsep}{2pt}
\caption{Top-5 similar players for six elite players in 2015/16 season}
\label{tab:elite6_top5}
\vspace{-1em}
\begin{subtable}[t]{0.32\textwidth}
\centering
\caption{Robert Lewandowski}
\begin{adjustbox}{max width=\linewidth}
\begin{tabular}{@{}r L{2.6cm} c c c@{}}
\toprule
Rk & Similar Player & Pos & Sim & Apps \\
\midrule
 1 & Gonzalo Higuaín & FWD & 1.000 & 35 \\
 2 & Luis Suárez & FWD & 1.000 & 35 \\
 3 & Zlatan Ibrahimović & FWD & 1.000 & 30 \\
 4 & Diego Costa & FWD & 1.000 & 28 \\
 5 & Karim Benzema & FWD & 1.000 & 27 \\
\bottomrule
\end{tabular}
\end{adjustbox}
\end{subtable}
\hfill
\begin{subtable}[t]{0.32\textwidth}
\centering
\caption{Lionel Messi}
\begin{adjustbox}{max width=\linewidth}
\begin{tabular}{@{}r L{2.6cm} c c c@{}}
\toprule
Rk & Similar Player & Pos & Sim & Apps \\
\midrule
 1 & José Callejón & AM & 0.999 & 38 \\
 2 & Xherdan Shaqiri & AM & 0.998 & 27 \\
 3 & Gareth Frank Bale & AM & 0.998 & 23 \\
 4 & Ángel Di María & AM & 0.998 & 28 \\
 5 & Domenico Berardi & AM & 0.998 & 29 \\
\bottomrule
\end{tabular}
\end{adjustbox}
\end{subtable}
\hfill
\begin{subtable}[t]{0.32\textwidth}
\centering
\caption{Andrés Iniesta}
\begin{adjustbox}{max width=\linewidth}
\begin{tabular}{@{}r L{2.6cm} c c c@{}}
\toprule
Rk & Similar Player & Pos & Sim & Apps \\
\midrule
 1 & Jean Michaël Seri & DM & 0.999 & 38 \\
 2 & Marek Hamšík & DM & 0.999 & 38 \\
 3 & Daniele Croce & DM & 0.999 & 28 \\
 4 & Paul Pogba & DM & 0.999 & 35 \\
 5 & Daniele Baselli & DM & 0.999 & 35 \\
\bottomrule
\end{tabular}
\end{adjustbox}
\end{subtable}

\begin{subtable}[t]{0.32\textwidth}
\centering
\caption{Sergio Ramos García}
\begin{adjustbox}{max width=\linewidth}
\begin{tabular}{@{}r L{2.6cm} c c c@{}}
\toprule
Rk & Similar Player & Pos & Sim & Apps \\
\midrule
 1 & Laurent Koscielny & CB & 1.000 & 33 \\
 2 & Antonio Rüdiger & CB & 1.000 & 30 \\
 3 & Kalidou Koulibaly & CB & 1.000 & 33 \\
 4 & Jeison Fabián Murillo Cerón & CB & 1.000 & 34 \\
 5 & Georg Niedermeier & CB & 1.000 & 18 \\
\bottomrule
\end{tabular}
\end{adjustbox}
\end{subtable}
\hfill
\begin{subtable}[t]{0.32\textwidth}
\centering
\caption{Philipp Lahm}
\begin{adjustbox}{max width=\linewidth}
\begin{tabular}{@{}r L{2.6cm} c c c@{}}
\toprule
Rk & Similar Player & Pos & Sim & Apps \\
\midrule
 1 & Alessandro Florenzi & WB & 0.999 & 33 \\
 2 & Danilo Luiz da Silva & WB & 0.999 & 24 \\
 3 & Fabrizio Cacciatore & WB & 0.999 & 29 \\
 4 & Daniel Brosinski & WB & 0.999 & 30 \\
 5 & Ashley Young & WB & 0.999 & 18 \\
\bottomrule
\end{tabular}
\end{adjustbox}
\end{subtable}
\hfill
\begin{subtable}[t]{0.32\textwidth}
\centering
\caption{Manuel Neuer}
\begin{adjustbox}{max width=\linewidth}
\begin{tabular}{@{}r L{2.6cm} c c c@{}}
\toprule
Rk & Similar Player & Pos & Sim & Apps \\
\midrule
 1 & Claudio Bravo & GK & 1.000 & 32 \\
 2 & Antonio Rodríguez Martínez & GK & 0.999 & 10 \\
 3 & Roman Bürki & GK & 0.999 & 33 \\
 4 & Pepe Reina & GK & 0.999 & 37 \\
 5 & Kevin Trapp & GK & 0.999 & 34 \\
\bottomrule
\end{tabular}
\end{adjustbox}
\end{subtable}
\vspace{-1.8em}
\end{table*}

\section{Discussion}
The xG and VAEP experiments demonstrate the performance of the pretrained TabTransformer representations across two downstream prediction tasks.

For the xG task, as shown in Table~\ref{tab:overall_xg_results}, the TabTransformer outperforms the MLP baseline in calibration, as evidenced by its lower Brier score and LogLoss. The StatsBomb built-in xG model achieves the best overall calibration; this advantage is attributable to its larger proprietary training dataset and additional spatial features capturing player and opponent positioning.

Tables~\ref{tab:vaep_test_score} and \ref{tab:vaep_test_concede} show the performance of the TabTransformer and baseline models on the VAEP scoring and conceding sub-tasks. The TabTransformer models outperform task-specific gradient-boosted decision tree models using traditional feature engineering with respect to Brier score, the appropriate evaluation metric for such a probabilistic prediction task \cite{decroos2019}. The comparisons show that TabTransformer is superior to both the original VAEP implementation using CatBoost with SPADL features and the CatBoost implementation using the richer StatsBomb feature set. Consequently, improved performance is not solely due to more information, but can be attributed to a genuinely better representation of the information encoded in the features using dense embeddings.

These results demonstrate the potential of dense event embeddings as general-purpose representations for a range of football analytics tasks. With a pretrained TabTransformer, dense event representations can be computed directly from raw event data without task-specific model training.

From a deployment perspective, the primary advantage of the proposed approach is representational versatility rather than peak single-task performance. In the BERT fine-tuning paradigm \cite{devlin2018}, a single pretrained model is adapted to diverse downstream tasks with minimal additional training, substantially reducing data and engineering overhead for each new application. The TabTransformer embedding demonstrated here exhibits this property: the same 911-dimensional representation was applied to two structurally different downstream tasks, sequential xG prediction and tabular VAEP estimation, without retraining the backbone encoder.

\section{Conclusion}
We propose a TabTransformer-based framework for learning universal dense representations of football event data through feature-level Masked Language Modeling pretraining on anonymized StatsBomb event data.
By stripping player and team identifiers before pretraining, the model learns action semantics that generalize across players and leagues rather than approximating individual player reputation.
Evaluation on Expected Goal prediction and VAEP scoring and conceding estimation demonstrates that the learned representations achieve competitive performance with task-specific baselines, beating an MLP baseline for xG and outperforming the current state-of-the-art implementation for VAEP.
Current limitations include the performance gap relative to task-specific gradient boosting models on the xG and VAEP scoring tasks, and the absence of an empirical evaluation in data-scarce regimes.
Future work will address these gaps through qualitative analysis of the embedding geometry to validate semantic clustering of action types, controlled experiments in data-scarce scenarios, and investigation of more expressive pretraining objectives beyond single-feature masking.

\bibliographystyle{unsrt}
\bibliography{reference, phd_library}

\appendix




\end{document}